\documentclass[letterpaper, 10 pt, conference]{ieeeconf}  

\usepackage{cite}
\usepackage{amsmath,amssymb,amsfonts}
\usepackage{algorithmic}
\usepackage{graphicx}
\usepackage{algorithm,algorithmic}
\usepackage{hyperref}
\hypersetup{hidelinks=true}
\usepackage{textcomp}

\usepackage{booktabs} 
\usepackage{multirow}
\usepackage{pifont}
\newcommand{\cmark}{\ding{51}}%
\def\BibTeX{{\rm B\kern-.05em{\sc i\kern-.025em b}\kern-.08em
    T\kern-.1667em\lower.7ex\hbox{E}\kern-.125emX}}
\newenvironment{IEEEkeywords}{\begin{center}\bfseries Keywords\end{center}\begin{quote}}{\end{quote}}

\begin{document}
\title{Lightweight Sequential Transformers for Blood Glucose Level Prediction in Type-1 Diabetes}
\author{Mirko Paolo Barbato, Giorgia Rigamonti, Davide Marelli, Paolo Napoletano,~\IEEEmembership{Senior~Member,~IEEE}%
\thanks{The authors are with the Department of Informatics, Systems and Communication, University of Milano-Bicocca, Milano 20126, Italy (e-mail: mirko.barbato@unimib.it; giorgia.rigamonti@unimib.it; davide.marelli@unimib.it; paolo.napoletano@unimib.it).}
}

\maketitle

\begin{abstract}  
Type 1 Diabetes (T1D) affects millions worldwide, requiring continuous monitoring to prevent severe hypo- and hyperglycemic events. While continuous glucose monitoring has improved blood glucose management, deploying predictive models on wearable devices remains challenging due to computational and memory constraints. To address this, we propose a novel Lightweight Sequential Transformer model designed for blood glucose prediction in T1D. By integrating the strengths of Transformers' attention mechanisms and the sequential processing of recurrent neural networks, our architecture captures long-term dependencies while maintaining computational efficiency. The model is optimized for deployment on resource-constrained edge devices and incorporates a balanced loss function to handle the inherent data imbalance in hypo- and hyperglycemic events. Experiments on two benchmark datasets, OhioT1DM and DiaTrend, demonstrate that the proposed model outperforms state-of-the-art methods in predicting glucose levels and detecting adverse events. This work fills the gap between high-performance modeling and practical deployment, providing a reliable and efficient T1D management solution. 
\end{abstract}


\section{Introduction}
\label{sec:introduction}

Type 1 Diabetes (T1D)~\cite{type_1} is a chronic autoimmune condition requiring lifelong blood glucose concentration (BGC) monitoring to prevent life-threatening complications such as hypoglycemia (BGC below 70 mg/dL~\cite{hypoglycemia}) and hyperglycemia (BGC above 180 mg/dL~\cite{hyper}). Continuous glucose monitoring (CGM)~\cite{cgm} technology has transformed T1D management by providing real-time tracking of glucose levels, enabling timely interventions and improving patient outcomes. However, predicting future glucose levels accurately remains a critical challenge, as it directly impacts insulin administration and overall disease management.

State-of-the-art methods for BGC prediction leverage complex deep-learning architectures, such as Transformers and Recurrent Neural Networks (RNNs)~\cite{ecai}, to capture temporal dependencies and achieve high predictive accuracy. Despite their effectiveness, these methods are computationally intensive and have large memory footprints, limiting their deployment on resource-constrained wearable devices~\cite{zhu2024population,nasser2021iot}. Cloud-based solutions mitigate computational demands but introduce privacy concerns, latency issues, and reliance on network connectivity~\cite{bhat2017novel,aminizadeh2023applications}, reducing their reliability for real-world applications. Furthermore, existing models often struggle with the inherent imbalance in glucose datasets, where adverse events (e.g., hypoglycemia) occur less frequently than normal glucose levels, leading to suboptimal performance in detecting critical events.

To address these limitations, this paper introduces a novel \textit{Lightweight Sequential Transformer} model for blood glucose prediction in individuals with T1D. The proposed architecture combines the attention mechanism of Transformers with the sequential processing capability of RNNs, enabling it to effectively capture both long-term and short-term temporal dependencies in BGC data. By employing a compact neural network design, the model is optimized for deployment on wearable devices with limited computational resources. Additionally, a novel balanced loss function is introduced to improve the detection of hypo- and hyperglycemic events, addressing the issue of dataset imbalance.

Experiments conducted on two benchmark datasets, OhioT1DM~\cite{ohio} and DiaTrend~\cite{diatrend}, demonstrate that the proposed model outperforms existing state-of-the-art methods in predictive accuracy and adverse event detection. The model supports both single- and multimodal inputs, allowing flexibility for various use cases, and is validated through an ablation study exploring its performance under different training configurations. Code of the proposed method is available at: \url{https://github.com/unimib-islab/Diabetes_Sequential_transformer]}

This work proposes a methodology to fill the gap between high-performance blood glucose prediction and practical deployment, aiming to provide reliable and efficient solutions in edge computing systems. The proposed approach is a step forward in improving T1D management, thus enabling real-world adoption of advanced predictive models, thanks to its performance and adaptability.
%

\section{Related work}
\label{sec:sota}

Over time, machine and deep learning techniques have proven to be accurate and reliable tools for estimating BGC~\cite{ecai}. Various approaches for BGC prediction have been explored, ranging from traditional statistical models, such as the Autoregressive Integrated Moving Average (ARIMA)~\cite{arima} and the Unobserved Components Model (UCM)~\cite{comparison}, to more advanced machine learning algorithms, including Random Forest and XGBoost~\cite{blood}. While these methods laid the groundwork for BGC prediction, they cannot often model complex temporal dependencies and physiological interactions, limiting their performance in dynamic, real-world scenarios.

More recently, deep learning approaches have demonstrated significant advances in robustness and accuracy~\cite{xie2020benchmarking,xing2022continuous}, with architectures such as Recurrent Neural Networks (RNNs)~\cite{chen2018dilated,martinsson2020blood}, convolutional RNNs (CRNNs)~\cite{li2019convolutional}, and Long Short-Term Memory networks (LSTMs)~\cite{martinsson2018automatic} gaining popularity. However, these methods often struggle to effectively capture long-term dependencies in time-series data due to challenges such as vanishing gradients and sequential processing bottlenecks.

In contrast, Transformer models, with their self-attention mechanisms, are well-suited to overcome these limitations by better capturing long-range dependencies and enabling parallel processing of input sequences. Their widespread application has created new opportunities in time-series prediction~\cite{xue2024bgformer,bian2024hybrid,zhu2023edge,sergazinov2023gluformer,acuna2023predicting}. Notable Transformer-based approaches for BGC prediction include Informer~\cite{zhou2021informer} and TimesNet~\cite{wu2022timesnet}, both of which have been successfully tested on this task~\cite{xue2024bgformer}. Gluformer~\cite{sergazinov2023gluformer}, specifically designed for glucose level prediction, also incorporates uncertainty quantification, enhancing its potential clinical applicability. Additionally, hybrid architectures that combine Transformers with LSTMs, such as the Hybrid Transformer-LSTM model~\cite{bian2024hybrid}, have been proposed to further leverage the strengths of both frameworks, demonstrating improved performance on glucose forecasting tasks.

State-of-the-art approaches for BGC prediction can be broadly categorized into single-modality and multimodal methods. Single-modality methods rely exclusively on CGM data, while multimodal approaches integrate additional features—such as basal insulin, bolus doses, carbohydrate intake, and other physiological parameters—to improve prediction accuracy. Transformer-based single-modality methods, such as Informer and TimesNet, have demonstrated strong performance, highlighting the scalability of attention mechanisms in time-series tasks.

For multimodal approaches, BG-BERT~\cite{bg-bert} and Bi-GRU~\cite{rigamonti2024improving} represent the current state-of-the-art. BG-BERT leverages a self-supervised learning system based on the BERT architecture and a masked auto-encoder to extract rich contextual information, enabling robust adverse event detection. Bi-GRU, on the other hand, employs the Bidirectional Gated Recurrent Unit (Bi-GRU) architecture to effectively capture dependencies and patterns in sequential data, offering a parameter-efficient alternative for multimodal modeling~\cite{rigamonti2024improving}. However, despite their strengths, these approaches often face challenges when deployed on resource-constrained edge devices due to their computational demands.

Despite the advancements achieved by existing methods, the demand for efficient and lightweight models that can deliver high performance in both single- and multimodal scenarios remains largely unfulfilled. The proposed \textit{Lightweight Sequential Transformer} aims to bridge this gap by combining the efficiency of sequential processing with the power of attention mechanisms, enabling accurate glucose predictions with minimal computational overhead.

\section{Methodology}
\label{sec:method}

In this section, we outline the problem addressed and the solutions proposed in this work. Specifically, we detail both the designed architecture and the training strategy employed to enable the model to effectively predict BGC and recognize adverse events.

\subsection{Problem Formulation}

A glucose forecaster is a computational model $\mathcal{M}$ that predicts future glucose levels given a sequence of observed data. The input of the model $\mathcal{M}$ is a sequence of $T$ glucose observations $\mathbf{g} = \{g_{1}, \cdots,g_{T}\}$, where the length of $T$ depends on the forecasting horizon and usually ranges from 2 to 4 hours~\cite{bg-bert}. The output is a sequence of $L$ glucose predictions $\mathbf{\hat{g}} = \{\hat{g}_{1}, \cdots,\hat{g}_{L}\}$, where $L$ stands for forecasting horizon that usually is 30 or 60 minutes~\cite{deep}. When available, it is possible to use a multimodal approach by exploiting extra features, such as quantities of basal, bolus, carbs, and other physiological parameters, as input of the model $\mathcal{M}$ to get more accurate predictions. Extra features are usually pre-processed to obtain a sequence of the same length of $\mathbf{g}$~\cite{deep}. 

\subsection{Proposed forecaster model}

\begin{figure*}[tb]
\centering
  \includegraphics[width=0.68
\linewidth]{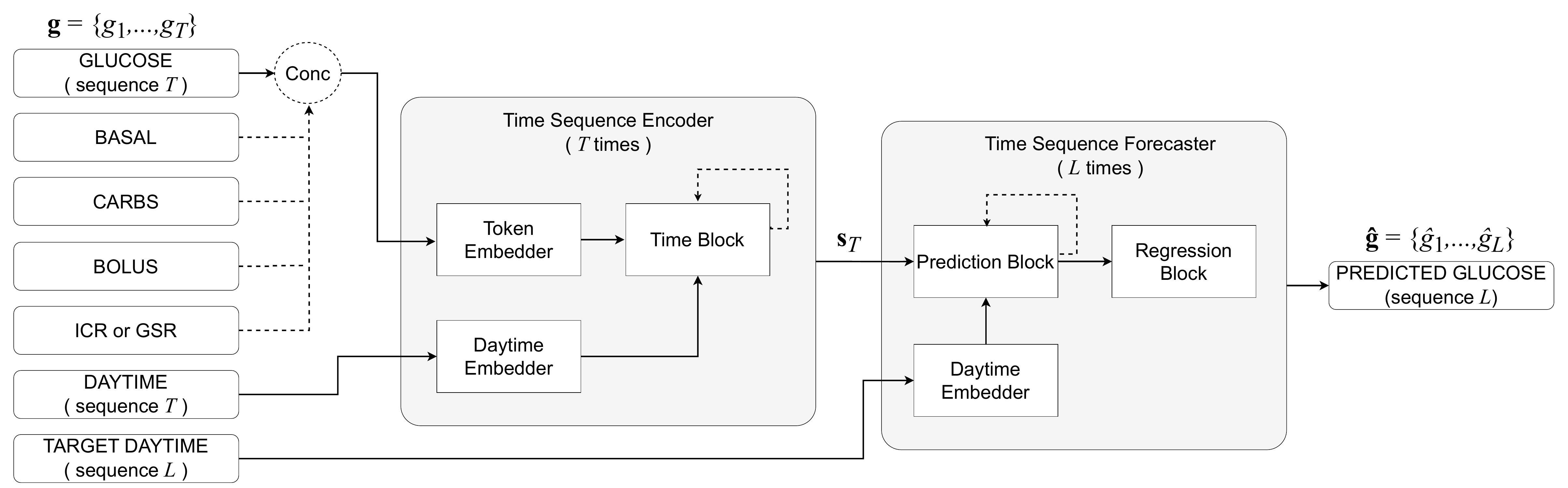}
  \caption{The architecture of the proposed Sequential Transformer $\mathcal{M}$.}
  \label{fig:pipeline}
\end{figure*}

The proposed Sequential Transformer draws inspiration from the architecture of the classical Transfomers~\cite{vaswani2017attention} and RNNs~\cite{grossberg2013recurrent}. The main idea is to create an architecture that exploits the concept of attention typical of Transformers without losing the notion of sequentiality that characterizes time series, which we consider fundamental for accurately predicting future information. A standard Transformer handles the entire observed sequence all at once, giving the same importance to all the samples over time. However, this approach may not fully capture the temporal dynamics necessary for tasks like glucose level prediction. By contrast, similarly to RNNs, the Sequential Transformer processes the elements of the observed sequence incrementally, one at a time. This strategy emphasizes the most recent observations in the sequence, which are temporally closer to the glucose levels to predict. This approach aligns with the glucose behavior over time, which typically fluctuates at a low frequency due to factors such as meal intake, insulin response, and circadian rhythms~\cite{kovatchev2016glucose}.
At the same time, the Sequential Transformer retains the attention mechanism (via query, key, and value), which is well-known for outperforming traditional architectures because it allows the model to capture long-range dependencies, handle variable-length sequences effectively, and prioritize relevant information dynamically~\cite{vaswani2017attention}.
Furthermore, the sequential nature of our approach, thanks to the parameter-sharing policy, inherently reduces the number of parameters compared to a standard Transformer, making it more efficient.





The proposed model $\mathcal{M}$, depicted in Figure~\ref{fig:pipeline}, is designed to handle both single-modality inputs, such as glucose sequences, and multi-modality inputs, including feature sequences in addition to the glucose one. Additionally, the model accepts the daytime information corresponding to the observed input sequences.
The model architecture $\mathcal{M}$ consists of the concatenation of a Time Sequence Encoder $\mathcal{E}$ and a Time Sequence Forecaster $\mathcal{F}$. Each of these modules is based on the Multi-Layer Perceptron ($\operatorname{MLP}$) operation and a slightly modified version of the Multi-head Cross-attention operation~\cite{wen2023distract}, which we refer to as Sigmoid Cross-attention (SCA). These are defined as follows:

\paragraph{Multi-Layer Perceptron} an $\operatorname{MLP}$ is defined as a sequence of two blocks, each composed of a linear projection, a GELU activation function~\cite{hendrycks2016gaussian}, and a dropout layer. Specifically, given a vector $\mathbf{x}$, each block is defined as $\mathbf{\operatorname{dropout}(\operatorname{GELU}(\operatorname{Linear}(x)))}$.

\paragraph{Sigmoid Cross-Attention}
a single-head cross-attention operation is defined as: $\operatorname{{c}}_{t}^{att}(\mathbf{q}_{t}, \mathbf{k}_{t}, \mathbf{v}_{t}) = \sigma\left(  \frac{\mathbf{q}_{t}\times \mathbf{k}_{t}^{\mathsf{T}}}{\sqrt{d}} \right) \times \mathbf{v}_{t}$, where $\mathbf{q}_{t}$, $\mathbf{k}_{t}$ and $\mathbf{v}_{t}$ are the query, key and value at time $t$, respectively, $d$ is a scale factor and $\sigma$ is the sigmoid activation function. The query, key, and value share the same size $N$, which defines the embedding dimension of the entire model. Cross-attention computes attention between different sequences: queries are derived from one sequence, while keys and values come from another sequence. A Multi-head Cross-attention module, denoted as $\operatorname{{c}}_{t, D}^{att}$, consists of $D$ heads that are processed in parallel and concatenated before being linearly projected through a linear layer.

\subsection{Time Sequence Encoder}

\begin{figure}[tb]
\centering
    \includegraphics[width=0.9\linewidth]{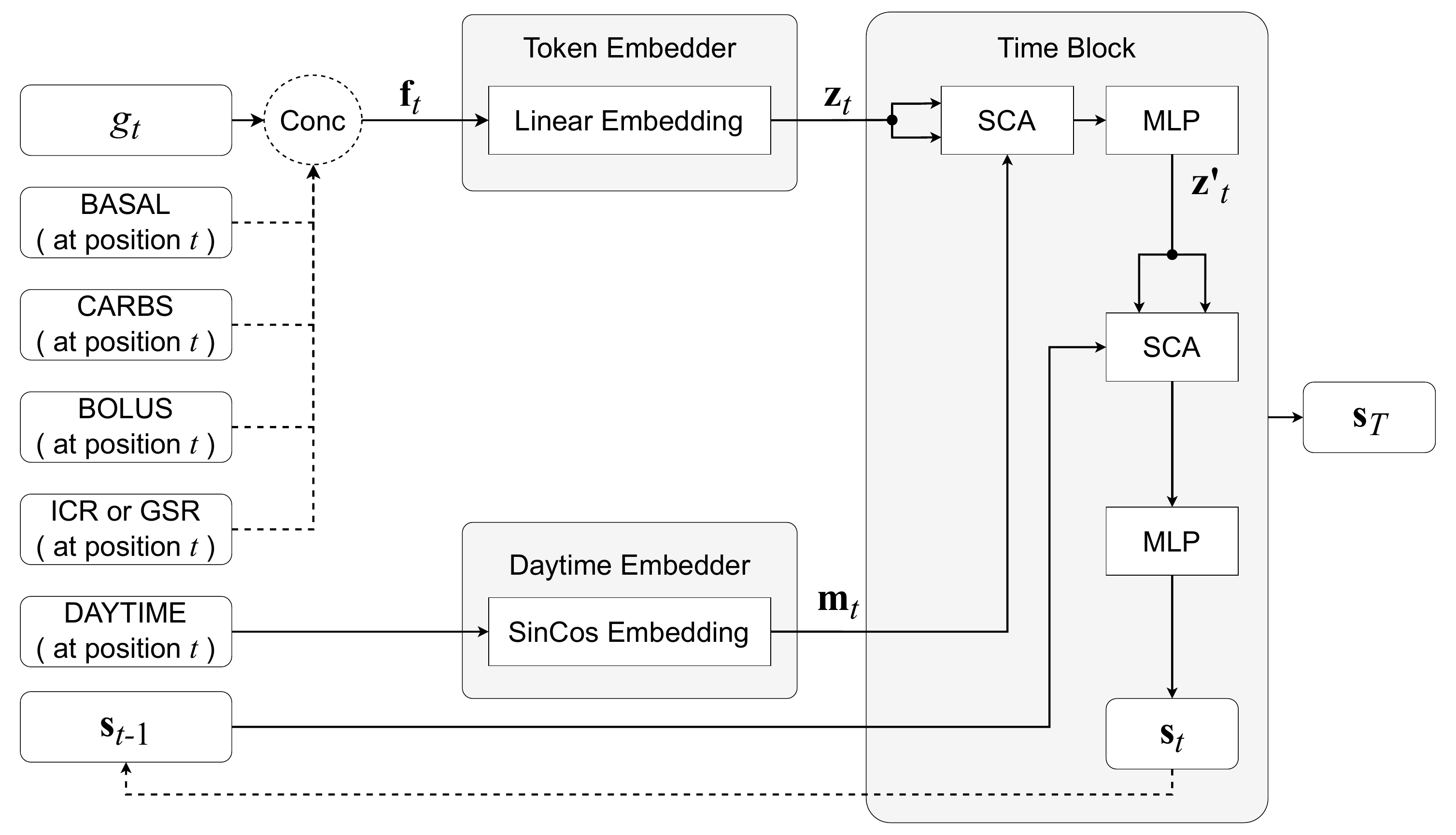}
    \caption{Time Sequence Encoder $\mathcal{E}$.}
    \label{fig:TSEncoder}
\end{figure}

The model $\mathcal{E}$, shown in Figure~\ref{fig:TSEncoder}, encodes the inputs sequentially using a Token Embedder, a DayTime Embedder, and a Time Block module. The inputs consist of the features of the observed sequence at time $t$ defined as $\mathbf{f}_{t}$, the corresponding daytime at $t$, and a vector $\mathbf{s}_{t-1}$ of dimension $N$.

In the single-modality case, $\mathbf{f}_{t}$ corresponds to the glucose level $\mathbf{g}_{t}$; in the multi-modality case, instead, $\mathbf{g}_{t}$ at time $t$ is concatenated with the other additional features.
Instead, the vector $\mathbf{s}_{t-1}$ encodes the information and history of all observed sequences from time $1$ to time $t-1$. Initially, $\mathbf{s}_{0}$ is randomly initialized and is the same for all sequences.

\paragraph{Token Embedder}
it takes $\mathbf{f}_{t}$ as input and performs a Linear Embedding operation to (1) optionally fuse the multimodal input, and (2) project the input into an embedding of dimension $N$. The output is a vector denoted as $\mathbf{z}_{t}$.

\paragraph{DayTime Embedder}
\label{subsubsec:daytimeEmb}
it uses the sine and cosine functions to project the daytime at time $t$ into an embedding of the same size $N$, as in~\cite{vaswani2017attention}. The resulting embedding is defined as $\mathbf{m}_{t}$.

\paragraph{Time Block module}
it takes as inputs the embedding $\mathbf{z}_{t}$ representing the features at time $t$, the embedding $\mathbf{m}_{t}$ representing the daytime, and the encoding $\mathbf{s}_{t-1}$ representing the entire observed sequence up to $t-1$. 
In the first step, $\mathbf{z}_{t}$ and $\mathbf{m}_{t}$ are combined into a new vector $\mathbf{z'}_{t}$ using an $\operatorname{MLP}$ and a SCA mechanism. Specifically, $\mathbf{z'}_{t} = \operatorname{MLP}(\operatorname{{c}}_{t,D}^{att}(\mathbf{m}_{t}, \mathbf{z}_{t}, \mathbf{z}_{t}))$, where $\mathbf{m}_{t}$ serves as the query, and $\mathbf{z}_{t}$ acts as the key and value. This operation weights $\mathbf{z}_{t}$ based on the daytime.
In the second step, $\mathbf{z'}_{t}$ is combined with the sequence encoding $\mathbf{s}_{t-1}$ observed till that moment, resulting in a new vector $\mathbf{s}_{t}$ that encapsulates the entire sequence information, including observations at time $t$. 
This step also employs an $\operatorname{MLP}$ and SCA mechanism; however, in this case, $\mathbf{s}_{t-1}$ is considered as the query, while $\mathbf{z'}_{t}$ as the key and the value, emphasizing the most recent element of the observed sequence. 
Formally, $\mathbf{s}_{t} = \operatorname{MLP}(\operatorname{{c}}_{t}^{att}(\mathbf{s}_{t-1}, \mathbf{z'}_{t}, \mathbf{z'}_{t}))$.

As previously mentioned, the entire Time Sequence Encoder is applied iteratively to each element of the observed sequence over $t$, resulting in $T$ computations. After $T$ iterations, the final output is a vector $\mathbf{s}_{T}$, which encodes the entire observed sequence input.


\subsection{Time Sequence Forecaster}

\begin{figure}[tb]
\centering
    \includegraphics[width=0.9\linewidth]{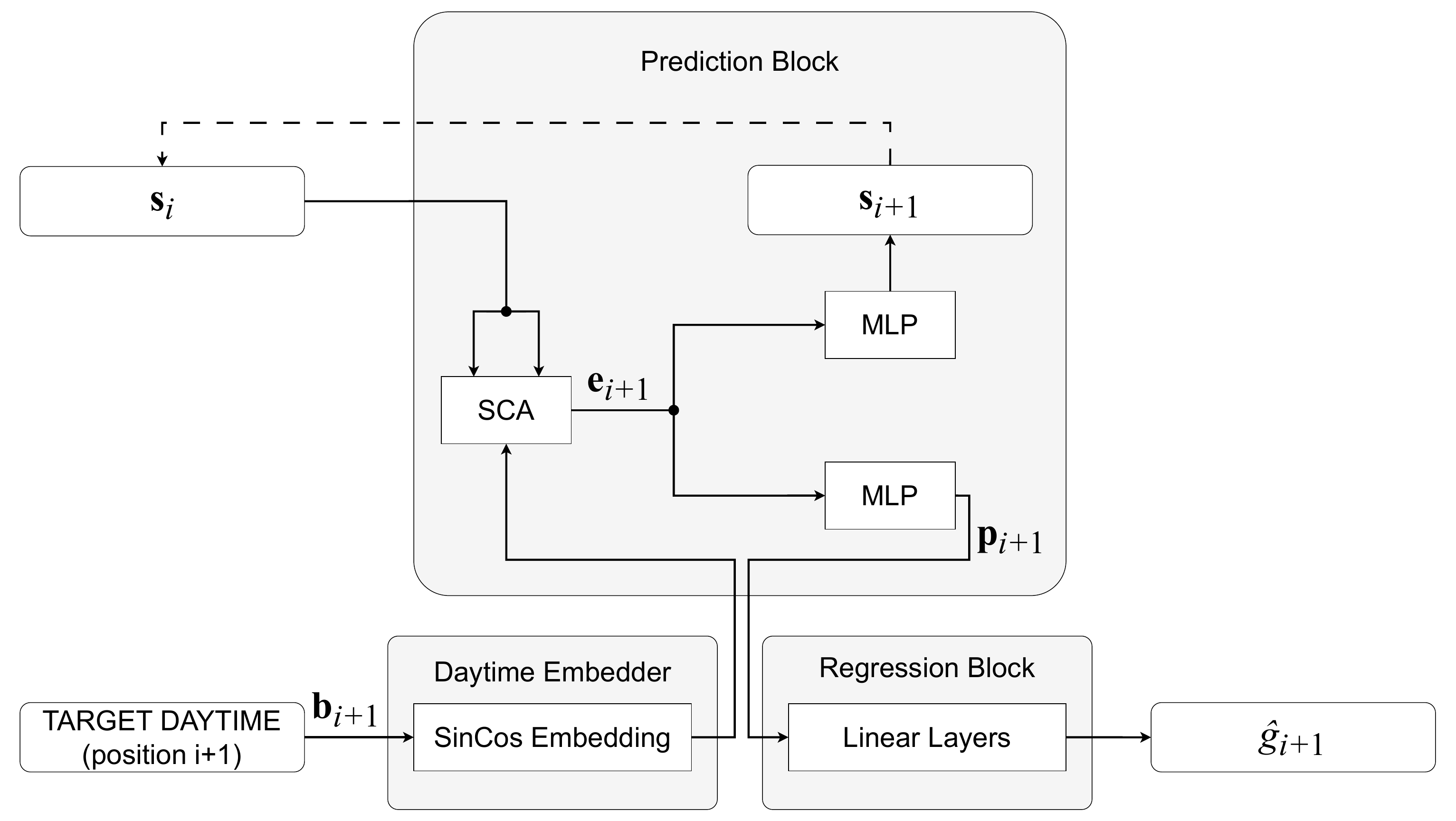}
    \caption{Time Sequence Forecaster $\mathcal{F}$.}
    \label{fig:TSForecaster}
\end{figure}

The Time Sequence Forecaster $\mathcal{F}$, illustrated in Figure~\ref{fig:TSForecaster}, predicts future glucose levels using a DayTime Embedder, a Prediction Block, and a Regression Block module. 
Even in this case, the inputs are processed sequentially with operations repeated $L$ times, where $L$ is the number of glucose levels to predict. The forecaster $\mathcal{F}$ takes as input a vector $\mathbf{r}_{i}$, where $i$ ranges from $0$ to $L-1$, and the daytime corresponding to the next glucose level to be predicted, at time $i+1$. Initially, $\mathbf{r}_{0}$ is set to $\mathbf{s}_{T}$, the output of the Time Sequence Encoder $\mathcal{E}$.  For subsequent stages, when $i \ge 1$, $\mathbf{r}_{i}$ encodes the information of the sequence from time $1$ to $T+i$, thus including all the observed sequence, from $1$ to $T$, and the estimated sequence, up to time to $T+i$. 
The output of $\mathcal{F}$ is $\hat{g}_{i+1}$, the predicted  glucose level at time ${i+1}_{th}$.

\paragraph{DayTime Embedder} it functions identically to the DayTime Embedder described in~\ref{subsubsec:daytimeEmb}. However, instead of processing the daytime of the observed sequence, it handles the daytime corresponding to the element to predict. Specifically, it processes the daytime at time $i+1$ and outputs an embedding $\mathbf{b}_{i+1}$ of dimension $N$.

\paragraph{Prediction Block}

it takes the embedding $\mathbf{b}_{i+1}$ of the daytime for the glucose level to be predicted at time $i+1$, and the embedding of the sequence $\mathbf{r}_{i}$ at time $i$. The output are two vectors of dimension $N$: the first, $\mathbf{r}_{i+1}$, represents the estimated sequence at time $i+1$, and the second, $\mathbf{p}_{i+1}$, represents the predicted glucose level at time $i+1$.

The process begins by combining $\mathbf{b}_{i+1}$ with $\mathbf{r}_{i}$ to obtain a vector $\mathbf{e}_{i+1}$. This combination is performed using a SCA mechanism. Formally, $\mathbf{e}_{i+1} = \operatorname{{c}_{t,D}^{att}}(\mathbf{b}_{i+1}, \mathbf{r}_{i}, \mathbf{r}_{i})$, where $\mathbf{b}_{i+1}$ serves as the query, and $\mathbf{r}_{i}$ is considered as key and value. Through this operation, $\mathbf{e}_{i+1}$ represents the estimated information at time $i+1$, computed by weighting the prior information using the daytime of the element to predict. The next step involves extracting $\mathbf{r}_{i+1}$ and $\mathbf{p}_{i+1}$ from $\mathbf{e}_{i+1}$. This is achieved using two separate $\operatorname{MLP}s$: $\mathbf{r}_{i+1} = \operatorname{MLP}(\mathbf{e}_{i+1})$ and $\mathbf{p}_{i+1} = \operatorname{MLP}(\mathbf{e}_{i+1})$.

\paragraph{Regression Block}

it is the final module of the proposed model. It takes as input $\mathbf{p}_{i+1}$ and predicts the corresponding glucose level $\hat{g}_{i+1}$ at time $i+1$. The module consists of a sequence of linear layers.

As described earlier, the entire Time Sequence Forecaster operates sequentially, iterating over $i$, for a total of $L$ times where $L$ corresponds to the number of glucose levels to predict. At the end of the $L$ iterations, the resulting output is the vector $\mathbf{\hat{g}} = \{\hat{g}_{1}, \cdots,\hat{g}_{L}\}$, where each element represents a predicted glucose level.

\subsection{Loss function}

The problem of BGC prediction is characterized by highly imbalanced datasets, particularly for normal, hypo-, and hyperglycemic events. Addressing this imbalance during training is crucial to ensure accurate predictions across all BGCs. Standard loss functions, such as Mean Squared Error (MSE), tend to be dominated by the majority class (i.e. normal glucose levels), which can lead to suboptimal performance in detecting rare but clinically critical events like hypoglycemia. To mitigate this, we use a \textit{balanced MSE}, defined as:

\begin{equation}
    \operatorname{Balanced MSE}(\mathbf{g},\mathbf{\hat{g}}) = 
    \sum_{i=1}^{L} w_i \cdot (g_i - \hat{g}_i)^2
    \label{eq:balanced_mse}
\end{equation}

\noindent
where $w_i$ represents the weight applied to the difference between the predicted value and the target value, and it is selected from the weights $w_{event}$ based on the type of event associated with the target value $\hat{g}_i$. 

The weights $w_{event}$ are computed by first subtracting the frequency of each event type (hypoglycemic, normal, and hyperglycemic) in the training set from one. These values are then scaled by a $relevance_{event}$ parameter, according to the decreasing frequency of the events in the training set: hypoglycemic events, being the least frequent, are multiplied by 3, hyperglycemic events by 2, and normal events by 1: $w_{event} = relevance_{event} \cdot \left(1 - \frac{num\_events}{total\_events}\right)$.

The balanced MSE ensures that the model pays proportionally more attention to less frequent but critical events during training, improving its sensitivity to hypoglycemia, even if it slightly sacrifices precision in the more frequent normal class. 

\section{Experiments}
\label{sec:experiments}
In this section, we present the experimental setup and metrics used to evaluate the proposed architecture. 

\subsection{Benchmark datasets}
\label{sec:materials}
We conducted our experiments on two benchmark datasets: OhioT1DM~\cite{ohio} and DiaTrend~\cite{diatrend}. 
The OhioT1DM dataset, widely used for BGC prediction, contains eight weeks of data from 12 individuals with T1D collected during the 2018 and 2020 challenges. It includes 5-minute CGM readings, insulin doses, blood glucose levels, carbohydrate intake, self-reported events (e.g., exercise, stress), and optional wearable fitness data (e.g., heart rate, activity levels).
In contrast, the DiaTrend dataset, one of the largest open-source resources for diabetes research, features longitudinal data from 54 individuals with T1D. It comprises 27,561 days of CGM data at 5-minute intervals, 8,220 days of insulin pump data (including basal insulin for 17 subjects), bolus doses, carbohydrate intake, pump settings, and detailed demographic and clinical profiles. For our analysis, we focused on the 17 subjects with detailed basal insulin data.



In line with prior studies~\cite{bg-bert,rigamonti2024improving}, we employed a temporal-based data split, partitioning the dataset into 64\% for training, 16\% for validation, and 20\% for testing.  This approach trains the model on earlier data, validates it on data closer to the training period, and evaluates it on future data, offering a realistic assessment of predictive performance. 
See Supplementary Materials\footnote{Supplementary materials here: \url{https://github.com/unimib-islab/Diabetes_Sequential_transformer/blob/23127d1e43e5c6bcec63d05a83ab800f5c31600d/Supplementary\%20material\%20-\%20IEEE_JBHI___Sequential_Transformers_for_Glucose_Prediction.pdf}\label{Sup_mat}} for more details on the pre-processing steps and the data split.

\begin{table*}[tb]
\centering	
 \caption{Overall performance on OhioT1DM Dataset\textsuperscript{*}}
 \label{tab:res_ohio}
 \resizebox{0.8\linewidth}{!}{
 \begin{tabular}{l c c c c c c c c c r c} 
 \toprule
 \multicolumn{1}{c}{} & \multicolumn{4}{c}{PH = 30 mins } & \multicolumn{4}{c}{PH = 60 mins} &  &  \\
 \cmidrule(lr){2-5}\cmidrule(lr){6-9}
Model & RMSE & TG & Hyper Sen~\textsuperscript{$\S$} & Hypo Sen~\textsuperscript{$\S$} & RMSE & TG & Hyper Sen~\textsuperscript{$\S$} & Hypo Sen~\textsuperscript{$\S$} & Features~\textsuperscript{o} & Params & Ranking \\
& (mg/dL) & (mins) & (\%) & (\%) & (mg/dL) & (mins) & (\%) & (\%) & \\
\midrule

Informer & 17.13 & 14.89 & 82.15 & 74.64 & 25.76 & 23.16 & 78.13 & 29.37 & - & 181K & 5 \\
TimesNet & \underline{13.63} & 14.71 & 90.22 & \textbf{85.72} & \underline{23.48} & 23.40 & 80.21 & 67.27 & - & 18749K & 6 \\  
Gluformer & 17.27 & \textbf{19.12} & 74.33 & \phantom{0}0.33 & 27.14 & 29.46 & 43.07 & 0.48 & - & 11247K & 8\\

Standard-T & 15.37 & 14.90 & 69.11 & \phantom{0}8.96 & 24.43 & 30.58 & 50.09 & 0.89 & - & 107K & 7 \\
Sequential-T & 14.96 & 17.56 & \textbf{96.26} & 75.82 & 26.42 & \underline{33.11} & \textbf{90.57} & \underline{68.95} & - & 123K & 1 \\

\midrule

BG-BERT\textsuperscript{$\dag$} & 14.02 & 16.56 & 82.54 & 73.24 & 23.67 & 31.16 & 69.24 & 54.12 & \cmark & 2091K & 4 \\
Bi-GRU\textsuperscript{$\ddag$} & \textbf{13.04} & 17.57 & 84.13 & 80.03 & \textbf{22.52} & 32.99 & 66.48 & 61.03 & \cmark & 633K & 2 \\

Standard-T & 17.17 & 15.57 & 82.98 & 5.21 & 24.74 & 29.81 & 63.49 & 0.03 & \cmark & 107K & 6 \\
Sequential-T & 16.00 & \underline{17.79} & \underline{94.66} & \underline{81.45} & 28.99 & \textbf{33.59} & \underline{89.32} & \textbf{69.96} & \cmark & 123K & 3 \\

\bottomrule
\end{tabular}}
 \\
 \vspace{0.3em}
 \raggedright{
    \scriptsize{\textsuperscript{*} In \textbf{bold} the best score and \underline{underlined} the second best score; \textsuperscript{o} The Features column indicates if the model considers a multimodal approach using all the features available (\cmark), or uses only the glucose level (-); {$\S$} Hyper Sen and Hypo Sen indicate the detection sensitivity of hyperglycemia and hypoglycemia respectively; {$\dag$} Values reported form~\cite{bg-bert}; {$\ddag$} Values reported form~\cite{rigamonti2024improving}.}\\
 } 

\end{table*}

\subsection{Baseline}
To demonstrate the advantages of the Sequential Transformer, we compare its effectiveness against state-of-the-art methods, including transformer-based models such as Informer~\cite{zhou2021informer}, TimesNet~\cite{wu2022timesnet}, Gluformer~\cite{sergazinov2023gluformer}, and BG-BERT~\cite{bg-bert}, as well as Bi-GRU~\cite{rigamonti2024improving}, with a particular focus on performance and model parameter count. Specifically, we benchmark it about two key settings: single-modal, using only the CGM feature, and multimodal, adopting the same features as BG-BERT and Bi-GRU, such as CGM, carbohydrate intake, bolus insulin dose, basal insulin rate, and Insulin-to-Carb Ratio (ICR) or Galvanic Skin Response (GSR), depending on the dataset being utilized (DiaTrend or OhioT1DM, respectively). Informer and Timesnet implementations are taken from \url{https://github.com/thuml/Time-Series-Library}.


To support our hypothesis, we also compared our method with an architecture based on the classical transformer encoder. The Standard Transformer architecture consists of a standard transformer encoder~\cite{vaswani2017attention} followed by the same regression layers as our method (see Supplementary Materials\footref{Sup_mat} for details).

\subsection{Experimental Setup}

All experiments are carried out on the two datasets described in the previous section, with two configurations (different observation and prediction windows) for each dataset.
Specifically, the first configuration features a 30-timestamp window (150 mins) divided into 24 timestamps (120 mins) for observed data and six timestamps (30 mins) for predictions. The second configuration uses a 60-timestamp window (300 mins) divided into 48 timestamps (240 mins) for observed data and 12 timestamps (60 mins) for predictions. These configurations are defined as OhioT1DM with PH = 30 mins, OhioT1DM with PH = 60 mins, DiaTrend with PH = 30 mins, and DiaTrend with PH = 60 mins. Since all configurations exhibit a significant imbalance between normal and adverse glucose levels, we applied the Synthetic Minority Over-sampling Technique (SMOTE) for data augmentation to the training set of all configurations~\cite{chawla2002smote}. This technique helps improve the representation of adverse glucose levels by generating synthetic samples based on k-nearest neighbors.

Besides the benchmark evaluation, we included an ablation study comparing the performance of the Sequential Transformer with different training settings. In detail, a balanced vs unbalanced training, with or without data augmentation, and with a single or multi-modality approach. The experiments are carried out in the same way as the benchmark evaluation.

The training setup for each experiment, in both benchmark and ablation studies, has been characterized by 2000 epochs and early stopping with a patience of 150.
We adopted the Adam optimizer with a learning rate of $1e^{-4}$ and a Reduce Learning Rate on the Plateau scheduler with 15 patience and 0.5 factor.
The training strategies used for the state-of-the-art models, such as the choice of the loss function, are the same as proposed as default by the corresponding original works.

\subsection{Evaluation Metrics}
We present three types of evaluations: (1) an analytical evaluation 
(both benchmark and ablation studies), based on four metrics commonly used in BGC prediction~\cite{bg-bert,rigamonti2024improving}: Root Mean Square Error (RMSE), Sensitivity on Hyper Events (Hyper Sen), Sensitivity on Hypo Events (Hypo Sen), and Time Gain (TG) (see Supplementary Materials\footref{Sup_mat} for more details); (2) a clinical evaluation using Clarke’s Error Grid Analysis (EGA)~\cite{ega} which is a widely used tool for interpreting BGC predictions and assessing the associated clinical risks in diabetes management. It shows measured and predicted BGC values on a plot divided into five zones, from A to E, based on proximity. Predictions in zones A and B are considered safe, with safety decreasing through zones C and D, while zone E represents the highest risk, where misdiagnosis could lead to fatal consequences; (3) deployment on edge devices is evaluated.

\section{Results and Discussion}
\label{sec:results}

\begin{table*}[tb]
\centering	
 \caption{Overall performance on DiaTrend Dataset\textsuperscript{*}}
 \label{tab:res_diatrend}
 \resizebox{0.8\linewidth}{!}{
 \begin{tabular}{l c c c c c c c c c r c} 
 \toprule
 \multicolumn{1}{c}{} & \multicolumn{4}{c}{PH = 30 mins } & \multicolumn{4}{c}{PH = 60 mins} & & \\
 \cmidrule(lr){2-5}\cmidrule(lr){6-9}
Model & RMSE & TG & Hyper Sen~\textsuperscript{$\S$} & Hypo Sen~\textsuperscript{$\S$} & RMSE & TG & Hyper Sen~\textsuperscript{$\S$} & Hypo Sen~\textsuperscript{$\S$} & Features~\textsuperscript{o} & Params & Ranking \\
& (mg/dL) & (mins) & (\%) & (\%) & (mg/dL) & (mins) & (\%) & (\%) & \\
\midrule

Informer & 16.57 & 13.56 & 79.83 & 50.76 & 25.85 & 23.41 & 70.14 & 14.78 & - & 181K & 5 \\
TimesNet & 15.53 & 13.64 & 88.17 & 69.56 & 25.45 & 21.98 & 76.65 & \underline{46.62} & - & 18749K & 9 \\
Gluformer & 17.29 & \textbf{17.74} & 79.11 & 0.03 & 25.22 & 29.94 & 55.64 & 0.43 & - & 11247K & 8 \\

Standard-T & 17.55 & 16.11 & 77.78 & 0.33 & 25.12 & 29.60 & 62.97 & 0.0 & - & 107K & 8 \\
Sequential-T & 16.21 & 16.00 & \textbf{94.80} & \textbf{73.77} & 26.71 & 28.41 & \underline{86.69} & \textbf{49.28} & - & 123K & 1 \\
\midrule

BG-BERT~\textsuperscript{$\dag$} & \underline{14.85} & 16.47 & 81.34 & 62.27 & \underline{24.95} & \underline{31.45} & 64.53 & 40.10 & \cmark & 2091K & 4 \\
Bi-GRU~\textsuperscript{$\ddag$} & \underline{14.64} & \textbf{16.66} & 81.31 & 63.82 & \textbf{24.57} & \textbf{32.47} & 65.72 & 31.64 & \cmark & 633K & 3\\

Standard-T & 18.00 & 16.18 & 78.22 & 0.25 & 37.32 & 28.93 & 37.40 & 0.00 & \cmark & 107K & 7\\
Sequential-T & 16.03 & 15.64 & \underline{93.89} & \underline{71.92} & 26.71 & 28.54 & \textbf{88.42} & 45.54 & \cmark & 123K & 2\\

\bottomrule
\end{tabular}}
 \\
 \vspace{0.3em}
 \raggedright{
    \scriptsize{\textsuperscript{*} In \textbf{bold} the best score and \underline{underlined} the second best score; \textsuperscript{o} The Features column indicates if the model considers a multimodal approach using all the features available (\cmark), or uses only the glucose level (-); {$\S$} Hyper Sen and Hypo Sen indicate the detection sensitivity of hyperglycemia and hypoglycemia respectively; {$\dag$} Values reported form~\cite{bg-bert}; {$\ddag$} Values reported form~\cite{rigamonti2024improving}.}\\
 } 
\end{table*}

Table~\ref{tab:res_ohio} and Table~\ref{tab:res_diatrend} respectively show the results achieved on the OhioT1DM and DiaTrend datasets, comparing the proposed method and training strategy with the state-of-the-art models. The reported experiments consider both the single and multimodal approaches. All architectures are evaluated on two dataset horizon configurations (30 and 60 minutes).
Each table shows the difference in terms of evaluation metrics and parameters, indicating whether the experiment uses only CGM or also additional features.
To facilitate the readability of the results, a Ranking column is also reported. The ranking is determined, considering the two PH configurations separately, 
by the average of the mean metrics' score, with the score of the parameters.
Such scores are computed by normalizing the results of the same metric (same column in the table) between $0$ and $1$, considering $1$ as the best result possible and $0$ as the worst. The score for the parameters is computed in the same way on the Params column. The average of the scores obtained for 30- and 60-minute horizons are then used to obtain the final rank independently from the horizon considered.

\begin{table*}[tb]
\centering	
  \caption{Sequential Transformer performance on OhioT1DM and DiaTrend datasets in different training settings\textsuperscript{*}}
 \label{tab:res_ohio_ablation}
 \resizebox{0.8\linewidth}{!}{
 \begin{tabular}{c c c c c c c c c c c c c c} 
 \toprule
  &\multicolumn{3}{c}{} & \multicolumn{4}{c}{PH = 30 mins} & \multicolumn{4}{c}{PH = 60 mins} \\
 \cmidrule(lr){5-8}\cmidrule(lr){9-12}
Dataset & Bal~\textsuperscript{+} & Aug~\textsuperscript{x} & Feat~\textsuperscript{o} & RMSE & TG & Hyper Sen~\textsuperscript{$\S$} & Hypo Sen~\textsuperscript{$\S$} & RMSE & TG & Hyper Sen~\textsuperscript{$\S$} & Hypo Sen~\textsuperscript{$\S$} & Ranking \\ & & &
& (mg/dL) & (mins) & (\%) & (\%) & (mg/dL) & (mins) & (\%) & (\%)\\
\midrule
\multirow{8}{*}{OhioT1DM} & - & - & - & \underline{13.56} & 15.75 & 77.54 & 20.14 & \underline{22.65} & 29.33 & 52.73 & 8.93  & 8 \\
&- & - & \cmark & 14.20 & 16.05 & 80.83 & 28.17 & 22.88 & 28.56 & 69.34 & 13.48 & 7 \\
&- & \cmark & - & \textbf{13.06} & 16.00 & 85.72 & 29.54 & \textbf{21.67} & 30.55 & 64.94 & 12.67 & 5\\
&- & \cmark & \cmark & 13.83 & 16.32 & 82.06 & 38.11 & 23.16 & 29.38 & 71.70 & 11.21 & 6 \\
&\cmark & - & - & 13.93 & 15.65 & \underline{95.42} & 68.07 & 24.47 & 30.63 & 87.79 & 65.88 & 4 \\
&\cmark & - & \cmark & 14.29 & 16.08 & 92.57 & 74.37 & 26.18 & 31.56 & \underline{89.71} & 59.40 & 3 \\
&\cmark & \cmark & - & 14.96 & \underline{17.56} & \textbf{96.26} & \underline{75.82} & 26.42 & \underline{33.11} & \textbf{90.57} & \underline{68.95} & 1 \\
&\cmark & \cmark & \cmark & 16.00 & \textbf{17.79} & 94.66 & \textbf{81.45} & 28.99 & \textbf{33.59} & 89.32 & \textbf{69.96} & 2 \\

\midrule
\multirow{8}{*}{DiaTrend} &- & - & - & 15.13 & 15.04 & 75.84 & 1.13 & 23.90 & \underline{29.90} & 59.88 & 5.25 & 8 \\
&- & - & \cmark & \textbf{14.75} & 14.75 & 81.95 & 28.13 & \underline{23.48} & \textbf{30.37} & 66.01 & 7.59 & 7 \\
&- & \cmark & - & 14.95 & 15.57 & 80.29 & 4.88 & 23.95 & 28.62 & 59.98 & 2.71 & 5 \\
&- & \cmark & \cmark & \underline{14.81} & 14.90 & 83.88 & 22.27 & \textbf{23.43} & 29.63 & 67.11 & 8.36 & 6 \\ 
&\cmark & - & - & 15.54 & 14.88 & 92.55 & 61.00 & 26.41 & 26.61 & \underline{88.31} & 36.48 & 3 \\
&\cmark & - & \cmark & 15.61 & 14.37 & 93.00 & 57.81 & 25.98 & 28.59 & 86.37 & 29.83 & 4 \\
&\cmark & \cmark & - & 16.21 & \textbf{16.00} & \textbf{94.80} & \textbf{73.77} & 26.71 & 28.41 & 86.69 & \textbf{49.28} & 1 \\
&\cmark & \cmark & \cmark & 16.03 & \underline{15.64} & \underline{93.89} & \underline{71.92} & 26.71 & 28.54 & \textbf{88.42} & \underline{45.54} & 2 \\
\bottomrule
\end{tabular}}
 \\
 \vspace{0.3em}
 \raggedright{
    \scriptsize{\textsuperscript{*} In \textbf{bold} the best score and \underline{underlined} the second best score; \textsuperscript{+} The Balance column indicates if the loss used was balanced (\cmark) or unbalanced (-); \textsuperscript{x} The Data aug column indicates if the data augmentation has been applied (\cmark) or not (-); \textsuperscript{o} The Features column indicates if the model considers a multimodal approach using all the features available (\cmark), or uses only the glucose level (-); \textsuperscript{$\S$} Hyper Sen and Hypo Sen indicate the detection sensitivity of hyperglycemia and hypoglycemia respectively;}\\
} 
\end{table*}

\paragraph{OhioT1DM}

as observable in Table~\ref{tab:res_ohio}, the best rank is achieved by the proposed model with a single-modality approach, while the multimodal version achieves the third rank. This means that, on the OhioT1DM dataset, the Sequential Transformer proves to be optimal for real-life situations requiring edge computing, thanks to its extremely low dimensionality in terms of parameters and its performance considering only CGM. The second best performing model is Bi-GRU~\cite{rigamonti2024improving}. However, it is important to notice that even if the model is relatively small in terms of parameters and compared with other architectures, its dimension can still be a problem on wearable devices. Another fundamental analysis regards the identification of future adverse events. The proposed strategy with the Sequential Transformer can always achieve the best or among the best results in detecting hyper and hypo 
events. On this particular issue, it is possible to observe that the multimodal Sequential Transformer outperforms the Bi-GRU, thus being more suitable in a real-life scenario.

\paragraph{DiaTrend}

Table~\ref{tab:res_diatrend} shows a trend similar to the OhioT1DM dataset. However, in this case, the Sequential Transformer takes both the best and second-best positions on the ranking.
The single-modality version achieves the best ranking, with the multimodal version reaching the second place. 
As before, the Sequential Transformer in every version achieves the best or among the best performance in detecting adverse 
events. The only exception is TimesNet~\cite{wu2022timesnet}, which achieves the second-best performance in Hypo Sen. Nonetheless, its number of parameters makes it unsuitable for real-life scenarios where edge computing results in a better solution.

\subsection{Ablation studies}
\label{sec:ablation}

In this section, we test different settings for the proposed method. We investigate the effectiveness of using balance and unbalanced loss for training, the advantages of data augmentation, and the exploitation of multi- and single-modality approaches. As for the afore-described experiments, a Ranking is defined to better identify which method performs best. This ranking only considers the evaluation metrics since all tested configurations use the same architecture.


The first rows of Table~\ref{tab:res_ohio_ablation} report the scores obtained on the OhioT1DM dataset with different settings. The best-ranked method is represented by our Sequential Transformer using the proposed balancing strategy, implementing data augmentation, and, as determined before, using only CGM. The second-best is the multimodal version, with the same setting in terms of loss and data augmentation. It is interesting to notice that, based on the ranking, the balanced versions represent the first four best choices, demonstrating the effectiveness of the defined training strategy.
Similar conclusions can be drawn for 
the DiaTrend dataset. 
The first rank is still taken by the single-modality setting with a balanced strategy and data augmentation. The second is the same setting but with a multimodal approach. Even in this case, it is possible to notice that the four best-ranked settings are all based on the balance training strategy, remarking once again on the importance of such implementation.

In summary, this study demonstrates that the proposed method coupled with the proposed training strategy, is optimal in edge computing systems, thanks to its dimension and the single-modality performance, which remove the problem of communication between multiple and heterogeneous devices~\cite{xu2022full}.
%
However, the results also indicate that the implementation of the multimodal approach can be refined, leaving room for potential future enhancements.

\subsection{Clarke Error Grid Analysis}




The Clarke Error Grid is a valuable instrument for assessing the clinical acceptability of BGC predictions at 30- and 60-minute horizons. In this analysis, we report the percentage of samples in each grid zone (see Supplementary Materials\footref{Sup_mat} for the plots visualization and more details).


For OhioT1DM, the proposed model achieves the following results: 94.15\%, 5.67\%, 0.01\%, 0.17\%, and 0.00\% in zones A, B, C, D, and E, respectively, for the 30-minute PH, and 82.54\%, 16.70\%, 0.29\%, 0.46\%, and 0.02\% for the 60-minute PH. On the DiaTrend dataset, the model performs similarly with 93.13\%, 6.64\%, 0.03\%, 0.21\%, and 0.00\% in zones A--E for the 30-minute PH, and 82.42\%, 16.60\%, 0.23\%, 0.72\%, and 0.03\% for the 60-minute PH.
These results highlight the model’s strong performance in predicting BGC at shorter time horizons (30 mins), with only minor deviations, as indicated by the small proportion of samples in zone B. Though there are a few samples in zones D and E, indicating larger errors, these are relatively rare. 
For longer horizons (60 mins), accuracy decreases as expected, but a significant proportion remains in zone A, showing the model’s reliability despite more clinically less accurate predictions. This reflects the usual trade-off between accuracy and time horizon in time-series modeling.

\subsection{Deployment on wearable devices}

Our method, used in its best configuration (only CGM), requires 0.49 MB of flash memory to store the model's weights. The RAM usage to temporarily store input data, activations, and output predictions is 320 KB in the 30-minute PH and nearly doubles, reaching 642 KB, in the 60-minute PH. These requirements can be easily satisfied by microcontroller units (MCUs) for the 30-minute PH, making our method suitable for edge computing deployment. More critical remains the 60-minute horizon for which few MCUs can provide enough RAM. In this case, the RAM usage can be lowered thanks to the recurrent structure of the model that allows the processing of the input time-series samples independently.
See Supplemental Materials\footref{Sup_mat} for further discussion on this topic.

\section{Conclusion}
\label{sec:conclusion}
In this paper, we defined a novel method for BGC prediction in T1D patients, aiming, at the same time, to design an architecture that is both accurate and deployable on low-resource edge computing systems. This task poses significant challenges, requiring lightweight and efficient models capable of operating within the constraints of edge devices. Furthermore, the nature of T1D presents additional challenges, including the inherent imbalance between normal, hypo-, and hyperglycemic events. The lower occurrence rates of these adverse events exacerbate the difficulty of building and training robust deep-learning models.
To address these challenges, we proposed the Sequential Transformer, a novel architecture that integrates the strengths of Transformers and RNNs. This design is coupled with a training strategy based on balanced MSE, specifically crafted to address the inherent imbalance in diabetes-related events.
Our approach delivers both high performance and a limited dimension model size, meeting the requirements of resource-constrained environments. The ablation study further validates the effectiveness of balanced MSE, demonstrating its superiority over conventional strategies. Moreover, the proposed model proved effective in both single-modality and multimodality scenarios, underscoring its adaptability and versatility.
These characteristics posed our solution as an effective and practical tool for deployment in edge computing systems and real-world applications.

Future studies will explore multimodal approaches and better utilize patient-specific features to improve event prediction. Personalizing the model for individual patients will be a pivot, along with investigating its deployment on edge devices for real-time applications.

\section*{Acknowledgment}
This work was funded by the National Plan for NRRP Complementary Investments (PNC, established with the decree-law 6 May 2021, n. 59, converted by law n. 101 of 2021) in the call for the funding of research initiatives for technologies and innovative trajectories in the health and care sectors (Directorial Decree n. 931 of 06-06-2022) - project n. PNC0000003 - AdvaNced Technologies for Human-centrEd Medicine (project acronym: ANTHEM)~\footnote{\url{https://fondazioneanthem.it/}}. This work reflects only the authors’ views and opinions, neither the Ministry for University and Research nor the European Commission can be considered responsible for them.

\bibliographystyle{IEEEtran}
\bibliography{bibliography}

\end{document}